\documentclass[sigconf, review=false]{acmart}

\usepackage{booktabs} 

\setcopyright{rightsretained}
\usepackage{multirow}
\usepackage{makecell}
\usepackage{tabularx}

\copyrightyear{2024}
\acmYear{2024}
\setcopyright{othergov}
\acmConference[ICVGIP 2024]{Indian Conference on Computer Vision Graphics and Image Processing}{December 13--15, 2024}{Bengaluru, India}
\acmBooktitle{Indian Conference on Computer Vision Graphics and Image Processing (ICVGIP 2024), December 13--15, 2024, Bengaluru, India}
\acmPrice{}
\acmDOI{10.1145/3702250.3702266}
\acmISBN{979-8-4007-1075-9/24/12}

\begin{document}
\title{GraphVL: Graph-Enhanced Semantic Modeling via Vision-Language Models for Generalized Class Discovery}
\titlenote{Produces the permission block, and
  copyright information}


\author{Bhupendra Solanki}
 \affiliation{
   \institution{IIT Bombay}
   \city{Mumbai}
  \state{Maharashtra}
  \country{India}
 }

 \author{Ashwin Nair}
 \affiliation{
   \institution{IISER Thiruvananthapuram}
  \city{Vithura}
  \state{Kerala}
  \country{India}
 }
 
\author{Mainak Singha}
 \affiliation{
   \institution{IIT Bombay}
   \city{Mumbai}
  \state{Maharashtra}
  \country{India}
 }

 \author{Souradeep Mukhopadhyay}
 \affiliation{
   \institution{IISc Bangalore}
   \city{Bangalore}
  \state{Karnataka}
  \country{India}
 }
 
\author{Ankit Jha}
 \affiliation{
   \institution{LNMIIT}
   \city{Jaipur}
  \state{Rajasthan}
  \country{India}
 }
 \author{Biplab Banerjee}
 \affiliation{
   \institution{IIT Bombay}
   \city{Mumbai}
  \state{Maharashtra}
  \country{India}
 }
 
\renewcommand{\shortauthors}{}

\begin{abstract}
Generalized Category Discovery (GCD) aims to cluster unlabeled images into known and novel categories using labeled images from known classes. To address the challenge of transferring features from known to unknown classes while mitigating model bias, we introduce GraphVL, a novel approach for vision-language modeling in GCD, leveraging CLIP. Our method integrates a graph convolutional network (GCN) with CLIP’s text encoder to preserve class neighborhood structure. We also employ a lightweight visual projector for image data, ensuring discriminative features through margin-based contrastive losses for image-text mapping. This neighborhood preservation criterion effectively regulates the semantic space, making it less sensitive to known classes. Additionally, we learn textual prompts from known classes and align them to create a more contextually meaningful semantic feature space for the GCN layer using a contextual similarity loss. Finally, we represent unlabeled samples based on their semantic distance to class prompts from the GCN, enabling semi-supervised clustering for class discovery and minimizing errors. Our experiments on seven benchmark datasets consistently demonstrate the superiority of GraphVL when integrated with the CLIP backbone.

\end{abstract}

%
%



    
\begin{teaserfigure}
    \centering
    \includegraphics[width=0.758\textwidth]{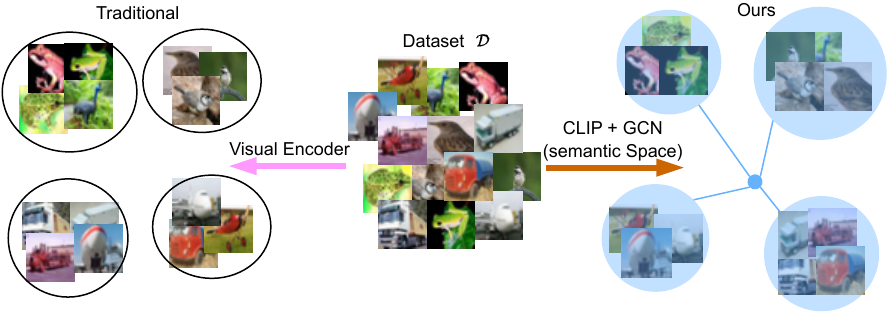}
    \vspace{-0.5cm}
    \caption{\textbf{Comparing Class discovery}: Traditional visual encoders distinguish classes and form clusters based on visual embeddings, which is often sub-optimal. Our proposed approach efficiently clusters classes using semantic information and maintains the semantic class topology using a graph convolution network, while optimizing the intraclass difference using a novel margin objective.}
    \label{fig:teaser-label}
    \Description{figure description}
\end{teaserfigure}
\maketitle

\section{Introduction}
\label{sec:intro}

Deep learning models have excelled in visual recognition tasks \citep{alexnet}, enabling novel machine learning approaches beyond supervised learning \citep{fsl, semisl}. So far, the importance of open-set recognition (OSR) and semi-supervised learning has been widely acknowledged for the downstream tasks. Semi-supervised learning, as described by \cite{fixmatch}, operates by integrating both labeled and unlabeled data within a closed-set framework.
In contrast, OSR \cite{gopenmax} deals with known and previously unknown classes in the unlabeled test set, often assigning an ``unknown'' label to potential novel class samples. However, clustering novel class data instead of labeling them could be more cost-effective. Novel class discovery (NCD) \cite{Fini2021AUO} aims to transfer knowledge from labeled to unlabeled datasets with distinct class labels for semantic clustering. Generalized class discovery (GCD) \cite{Vaze2022GeneralizedCD} relaxes NCD's assumptions for real-world applications, accounting for mixed known and novel class samples in the unlabeled set. However, despite its potential, GCD encounters several challenges. Existing models exhibit bias toward known classes due to labeled data availability and rely on vision-based backbones with limited semantic richness, limiting their utility on fine-grained datasets.

Large-scale vision-language models, like CLIP \cite{clip}, create semantically rich embeddings from image-text pairs. CLIP's potential for GCD remains largely unexplored, despite its success. K-means with pre-trained CLIP features outperforms ViT \cite{han2022survey} by around 10\% on CIFAR-100 \cite{Krizhevsky2009LearningML}. Bias issues persist, particularly for novel classes. Thus, our research question is: \textit{``How can we harness CLIP for an unbiased, discriminative semantic space for GCD?''}\\
\textbf{Introducing GraphVL:} To tackle these challenges, we present GraphVL, a novel framework for Generalized Category Discovery (GCD) that utilizes CLIP frozen backbones alongside a learnable Graph Convolutional Network (GCN) module in the semantic space. We propose three key considerations for learning an unbiased and discriminative embedding space, enhancing unlabeled data clustering (Fig. \ref{fig:teaser-label}).

Our approach enhances the semantic space by introducing a learnable GCN on CLIP's class embeddings, preserving neighborhood structures. This GCN maintains class relationships, ensuring that the neighborhood connections of pre-trained CLIP embeddings persist. A visual projector network adapts CLIP's vision encoder for task-specific use. Our training combines two metric-based objectives: increasing similarity among same-class samples and maximizing separation for different-class data in the non-semantic space. Additionally, a margin-based image-text contrastive loss ensures a discriminative non-semantic space, yielding more generalized and unbiased visual samples. We also perform prompt learning similar to COOP, aligning textual prompts from known classes with the semantic feature space obtained from the GCN layer. This alignment enhances the contextual meaning of the embeddings and improves the overall performance of our approach. For novel class discovery, we adapt the semi-supervised K-means algorithm \cite{Vaze2022GeneralizedCD}, utilizing similarity distributions based on learned class embeddings from the GCN. Our approach excels in GCD literature \cite{Vaze2022GeneralizedCD, promptcal} by systematically addressing bias and enhancing discriminativeness. 
Our main \textbf{contributions} are:\\
    \noindent - We tackle the challenging GCD problem, emphasizing bias reduction and enhanced discriminativeness. Our proposal combines the strengths of Vision-Language Models (VLMs), metric learning, and GCN. Our semi-supervised clustering in the semantic similarity-driven feature space improves class discovery.\\
    \noindent - To combat bias in labeled classes, we maintain semantic class relations through a learnable GCN module in a discriminative embedding space achieved via visual domain and cross-domain metric losses. We also use the prompt guided alignment between the generated graph and learned prompt embeddings.\\
    \noindent - We conduct extensive experiments comparing our GraphVL with benchmark NCD and GCD models re-implemented with the CLIP backbone, as well as the traditional DINO-based backbone \cite{Vaze2022GeneralizedCD}. Our findings consistently demonstrate the comprehensive superiority of GraphVL by a substantial margin of alteast 2\%.
\section{Relevant Works}

\subsection{Novel and Generalized Class Discovery}
Originally introduced by Han et al. \cite{Han2019LearningTD}, NCD aims to classify unlabeled newer categories by leveraging the understanding of labeled predefined categories. Earlier approaches that can be applied to this problem include \cite{Hsu2017LearningTC, Hsu2019MulticlassCW}, both of which utilize two models trained with labeled and unlabeled data, respectively, for general task transfer learning. RankStats \cite{Han2020AutomaticallyDA} addressed the NCD problem with a three-stage method. The model is first trained with self-supervision on all data for low-level representation learning. Then, it is further trained with full supervision on labeled data to capture higher-level semantic information. Finally, a joint learning stage is conducted to transfer knowledge from the labeled to unlabeled data with ranking statistics.
In another approach, \cite{Zhao2021NovelVC} proposed a model with two branches, one for global feature learning and the other for local feature learning, enabling dual ranking statistics and mutual learning for improved representation learning and new class discovery. In contrast, NCL \cite{neig_ncd} extracted and aggregated pairwise pseudo-labels for the unlabeled data using contrastive learning and generates hard negatives by combining labeled and unlabeled data in the feature space for NCD. A unified cross-entropy loss was introduced in \cite{Fini2021AUO} to enable joint training of labeled and unlabeled data, achieved by swapping pseudo-labels from labeled and unlabeled classification heads. 

The Generalized Category Discovery problem \cite{Vaze2022GeneralizedCD, cdadnet}, extends NCD in a more comprehensive manner by allowing unlabeled data to be sampled from both new and known classes. To address this challenge, \cite{Cao2021OpenWorldSL} proposes an uncertainty adaptive margin loss to reduce intra-class variances between new and known classes in an alternating fashion. In contrast, \cite{Vaze2022GeneralizedCD} proposes a semi-supervised contrastive learning approach using a large-scale pre-trained visual transformer (ViT), followed by the constrained K-Means clustering algorithm \cite{Arthur2007kmeansTA}, to address GCD. \cite{cdadnet} introduces generalization across domains for GCD task. However, our method still has limitations, as the frozen backbone may not adapt well to distributions that arise later, leading to a large number of false negatives and potential deterioration of semantic representation \cite{Chen2021IncrementalFN, Khorasgani2022SLICSL}. Finally, \cite{promptcal} suggested a two-stage contrastive affinity learning method with auxiliary visual prompts for GCD. \textit{Despite their success, these techniques do not explicitly tackle the issue of bias or leverage the vast semantic knowledge base, which serves as the main motivation behind our proposed approach, GraphVL.}

\subsection{Vision-Language Models and Prompt Learning}
Multimodal learning has been shown to outperform unimodal approaches in various inference tasks \cite{shottell,babenko2014neural} which require joint visual-semantic supervision. Recently, VLMs like CLIP \cite{clip} and Align \cite{align} have gained significant attention, as they are trained on large-scale image-text pairs in a contrastive manner to align visual and textual embeddings. VLMs enable the efficient transfer of learned visual information via prompt-based zero-shot and few-shot downstream tasks.
\begin{figure*}[ht!]
    \centering
    \vspace{-.3cm}
    \includegraphics[width=0.95\textwidth]{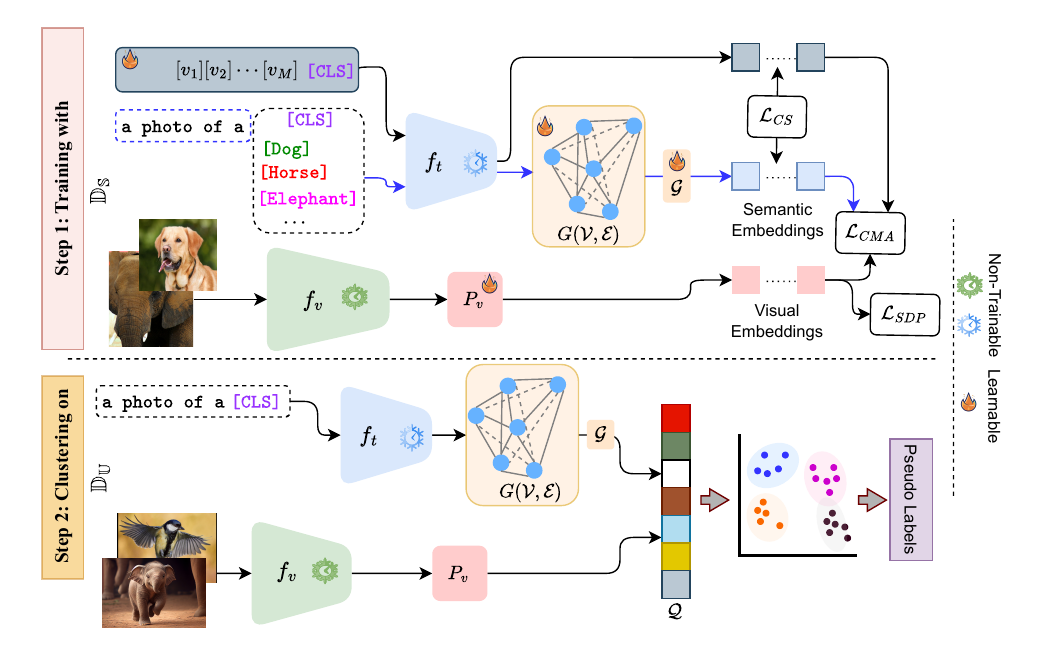}
    \vspace{-1.cm}
    \caption{Architecture overview of GraphVL: We propose a novel approach that integrates a learnable GCN-based text projector on top of CLIP's text encoder $f_t$, alongside a visual projector $P_v$ built on top of CLIP's frozen vision encoder $f_v$. The model is trained using various metric losses, including one in the non-semantic space ($\mathcal{L}_{SDP}$), visual-semantic spaces ($\mathcal{L}_{CMA}$), and contextual similarity ($\mathcal{L}_{CS}$). During inference, we apply a semi-supervised clustering technique to cluster the unlabeled data, utilizing the similarity distributions relative to the learned class embeddings as clustering features ($\mathcal{Q}$).}
    \label{fig:main_block}
    \vspace{-.3cm}
\end{figure*}
Prompt learning, a popular approach in natural language processing \cite{petroni2019language}, has also been applied to visual recognition tasks in VLMs like BERT \cite{bert}, with the goal of providing meaningful textual prompts for downstream tasks. Recent research has focused on automating prompt generation to eliminate manual intervention, leading to classification \cite{coop, cocoop, gopro, tpt, applenet, csaw}, domain generalization \cite{stylip, odgclip}, domain adaptation \cite{adclip, cosmo}, image retrieval \cite{splip} and class discovery \cite{cdadnet} tasks. However, these models often fix CLIP's text encoder while considering prompt tokens as the optimization variables, which may not preserve the discriminativeness of the semantic space. \textit{In contrast to existing literature, our approach employs hand-engineered prompts while incorporating a Graph Convolutional Network (GCN) on CLIP to learn a refined and discriminative semantic embedding space. This innovative combination of techniques is unique and sets our approach apart. Furthermore, we highlight the significance of margin-based contrastive losses in obtaining a discriminative semantic space that is highly effective for clustering tasks. }

\section{Methodology}
Consider the dataset $\mathbb{D}$, which is composed of both labeled supervised samples $\mathbb{D_S}$ and completely unlabeled samples $\mathbb{D_U}$, denoted as $\mathbb{D} \in \{\mathbb{D_S}, \mathbb{D_U}\}$. Specifically, $\mathbb{D_S}$ is comprised of $\{(x_i, y_i)\}_{i=1}^{N} \in X \times Y_S$, while $\mathbb{D_U}$ consists of $\{(x_j, y_j)\}_{j=1}^{M} \in X \times Y_U$, where $Y_S \subset Y_U$. Additionally, $Y_S$ corresponds to the label set for the \textsc{Known} classes denoted by ${C}_{kwn}$, while $Y_U$ represents the complete label set comprising both the \textsc{Known} classes and the \texttt{novel} classes denoted by ${C}={C}_{kwn} \cup {C}_{new}$. During the training stage, the model only has access to the labeled samples in $\mathbb{D_S}$, while the unlabeled samples in $\mathbb{D_U}$ are employed during inference. Following the approach in \cite{Vaze2022GeneralizedCD}, we assume that $|{C}|$ is known a priori. The goal is to accurately cluster the samples in $\mathbb{D}_{\mathbb{U}}$ into $|{C}|$ groups by transferring the representations learned from $\mathbb{D_S}$.

\subsection{Our proposed GraphVL}
\noindent\textbf{Model architecture and training:} The primary goal of using the GCD is to enable a recognition model to classify samples during testing, regardless of whether they belong to training classes or novel concepts. However, training on labeled classes may introduce bias, affecting clustering performance for novel classes and leading to suboptimal results. To address this, we propose maintaining semantic class relationships in the embedding space while maximizing inter-class separation. This strategy fosters a discriminative and well-regularized feature space, avoiding trivial solutions. To achieve this, in Fig. \ref{fig:main_block}, we utilize the pre-trained CLIP model to learn the embedding space, as it provides a promising solution. 

\noindent\textbf{Graph embeddings:} Consider the pre-trained vision and text encoders of CLIP, denoted by $f_v$ and $f_t$, respectively. For the classes in ${C}_{kwn}$, we utilize prompts of the form ``\texttt{a photo of a [CLS]}''. Let $y \in {C}_{kwn}$ represent the prompt for class $y$, and let $f_t(y)$ denote the corresponding prompt embedding from CLIP. To facilitate the embedding process, we introduce a learnable visual projector $P_v$ on top of $f_v$ and a Graph Convolutional Network (GCN) $\mathcal{G}$ that leverages the semantic graph structure $G(\mathcal{V}, \mathcal{E})$, obtained from the prompts using $f_t$. Here, $\mathcal{V}$ represents the node set of $G$, with each node corresponding to a class prompt. The edges in $\mathcal{E}$ ensure connectivity based on the k-nearest neighbor rule: $\mathcal{E}(i_1, i_2)=1$ if $y_{i_1} \in \mathcal{N}^k(y_{i_2})$, otherwise $\mathcal{E}(i_1, i_2)=0$, where $y_{i_1}, y_{i_2} \in {C}_{kwn}$. The learned visual and semantic embeddings are obtained by leveraging $P_v$ and $\mathcal{G}$. Additionally, to ensure that the GCN module $\mathcal{G}$ learns abstract semantic information corresponding to the input image, we perform the prompt learning strategy, as shown in Fig. \ref{fig:main_block}.\\
\noindent\textbf{Graph Convolutional Network (GCN):} In GCN, we entail the construction of a graph in which nodes signify class embedding vectors \(\mathcal{V}\), while edges \(\mathcal{E}\) encode the essential neighborhood connections between classes: \(\mathcal{G} = (\mathcal{V}, \mathcal{E})\). The graph is then fed into the GCN, which performs message-passing iterations to update the node representations. Specifically, the node embeddings \(\mathbf{H}^{(l)}\) at layer \(l\) are updated as follows: \(
\mathbf{H}^{(l+1)} = \sigma\left( \mathbf{D}^{-1} \mathbf{A} \mathbf{H}^{(l)} \mathbf{W}^{(l)} \right),
\)
where \(\mathbf{A}\) represents the adjacency matrix of the graph, \(\mathbf{D}\) is the degree matrix, \(\mathbf{H}^{(l)}\) is the node embedding matrix at layer \(l\), \(\mathbf{W}^{(l)}\) are the learnable weight matrices, and \(\sigma(\cdot)\) denotes a non-linear activation function (e.g., ReLU). This iterative process enables the GCN to assimilate crucial insights from neighboring nodes, thereby refining the node representations at each step. The final node embeddings \(\mathbf{H}^{(L)}\), where \(L\) is the number of layers, encapsulate the essential relationships and dependencies that intertwine the classes. In our case, \(\mathbf{H}^{(0)}\) would consist of the initial embeddings of these classes before any graph-based processing, derived from the CLIP encoder: \(
\mathbf{\bar{Y}} = \text{GCN}(\mathcal{G}, \mathbf{H}^{(0)}),
\)
where \(\mathbf{\bar{Y}}\) represents the final output node embeddings.\\
\noindent\textbf{Prompt learning:}
Following \cite{coop}, we use prompt learning to create generic prompts with $M$ learnable context vectors, ${v_1, v_2, \dots, v_M}$, each matching the dimensions of the word embeddings. For the $i^{th}$ class, the prompt $t_i$ is formulated as ${v_1, v_2, \dots, v_M, c_i}$, where $c_i$ represents the word embeddings for the class label $y$ (i.e., \texttt{[CLS]}). These context vectors are applied uniformly across all classes. We focus on aligning the contextual information between the GCN and prompt features rather than rigorous prompt learning. The embeddings are transformed into lower-cased byte pair encoding (BPE), assigning unique numeric IDs to each token. Each sequence, prefixed with [SOS] and suffixed with [EOS], is fixed at 77 tokens. These IDs are then converted into 512-dimensional word embedding vectors and input into $f_t$ to obtain $t_i$. Further details on prompt learning are in CoOp \cite{coop}. These text features maximize similarity with $\mathcal{G}(f_t(y))$ through a contextual similarity loss ($\mathcal{L}_{CS}$) to enhance generalizability.
\subsection{Loss Function \& Semi-supervised Clustering}
\noindent \textbf{Loss functions for training on $\mathbb{D_S}$:} We employ GraphVL and optimize it using following losses: i) \textit{Cross-modal Margin Alignment}, denoted as $\mathcal{L}_{CMA}$, which is based on the contrast between a given $z$ and $\{\mathcal{G}(y_c)\}_{c=1}^{|{Y_S}|}$, ii) \textit{Semantic Distinction Penalty}, denoted as $\mathcal{L}_{SDP}$, which is based on visual triplets $(z^a, z^p, z^n)$, with $z = P_v(f_v(x))$ and $(z^a, z^p)$ share the same class labels, while $(z^a, z^n)$ have different labels, and iii) \textit{Contextual Similarity Loss}, denoted as $\mathcal{L}_{CS}$.
We describe these losses in detail as,\\
\noindent\textbf{i) Cross-modal Margin Alignment:} In GraphVL, we aim to project the visual features into a semantically-induced space, leveraging the space learned by $\mathcal{G}$. This requires optimizing the image-text features in the contrastive manner. Specifically, for a image-label pair $(x,y) \in \mathbb{D_S}$ and letting $z=P_v(x)$ and $\bar{y}=\mathcal{G}(y)$, the \texttt{cosine} similarity between $(z, \bar{y})$ should be higher than that of $(z, \bar{y}_{-})$, where $y_{-} \neq y$ and $y_{-} \in \mathcal{C}_{kwn}$, by a pre-defined margin $\alpha$. Our proposed $\mathcal{L}_{CMA}$ loss is defined as follows,
\vspace{-0.15cm}
\begin{equation}
\begin{split}
    \mathcal{L}_{CMA} = - \underset{P_v, \mathcal{G}}{\text{argmin}} \underset{ (x,y) \sim P_{\mathbb{D_S}}}{\mathbb{E}} \text{log} \frac{\exp(\delta(z,\bar{y}))}{\underset{c=1}{\overset{|{Y_S}|}\sum}\exp(\delta(z,\bar{y}_c))} \\ + \underset{c=1}{\overset{|{Y_S}|}\sum} [\delta(z,\bar{y})-\delta(z,\bar{y}_c)-\alpha]_{+};
    \label{eq_unssl}
     \end{split}
\end{equation}
\noindent\textbf{ii) Semantic Distinction Penalty:} Learning discriminative features among the class is very important and to achieve this we use propose the semantic distinction penalty loss $\mathcal{L}_{SDP}$ on the features obtained from the $P_v$ on the supervised data $\mathbf{D_S}$, which emphasizes a high \texttt{cosine} similarity $(\delta)$ for the pair $(z^a, z^p)$, while minimizing it for $(z^a, z^n)$. This approach also ensures a compact non-semantic space with significant separation between the classes, which, in turn, helps in smooth mapping to the semantic space through $\mathcal{L}_{CMA}$. In Equation \ref{eq_SDP}, $P_{\mathbf{D_S}}$ defines the data distribution concerning the labeled samples in $\mathbf{D_S}$.
\begin{equation}
\vspace{-0.15cm}
\begin{split}
    \mathbf{L}_{SDP} = \underset{P_v}{\text{argmin}} \Biggr[\underset{(z^a, z^p) \sim P_{\mathbf{D_S}}}{\mathbb{E}} \left[ \delta(z^a, z^p)-1 \right] 
    \\  + \underset{(z^a, z^n) \sim P_{\mathbf{D_S}}}{\mathbb{E}} \left[ \delta(z^a, z^n) \right]\Biggr];
    \label{eq_SDP}
    \end{split}
\end{equation}
\begin{table*}[ht!]
    \centering
    \caption{Performance comparison to the state-of-the-art on the GCD task on the general purpose datasets. The first seven rows do not consider CLIP as a their feature backbone. The best accuracies are indicated in \textbf{bold}.}
    \vspace{-0.3cm}
    \scalebox{0.9}{
    \begin{tabular}{l|ccc|ccc|ccc}
    \toprule

       \multirow{2}{*}{\textbf{Method}}&\multicolumn{3}{c|}{\textbf{CIFAR10}}&\multicolumn{3}{c|}{\textbf{CIFAR100}}&\multicolumn{3}{c}{\textbf{ImageNet-100}}\\
       \cmidrule(lr){2-4} \cmidrule(lr){5-7}\cmidrule(lr){8-10}
    
       & \textsc{All} & \textsc{Known} & \textsc{New} & \textsc{All} & \textsc{Known} & \textsc{New}  & \textsc{All} & \textsc{Known} & \textsc{New} \\ 
       
        \midrule
        KMeans\cite{kmeans} &83.6&85.7&82.5& 52.0 & 52.2& 50.8&72.7 &75.5 &71.3 \\
        
        RankStats+ &46.8&19.2&60.5& 58.2 &77.6& 19.3&37.1&61.6&24.8 \\
        
        UNO+ &68.6&98.3&53.8&69.5&80.6&47.2&70.3&95.0&57.9\\
        
        GCD \cite{Vaze2022GeneralizedCD} &91.5&97.9&88.2&73.0&76.2&66.5&74.1&89.8&66.3 \\
        
        GPC \cite{Zhao2023LearningSG} &90.6&97.6&87.0 &75.4&84.6&60.1 &75.3&93.4&66.7 \\
        
        DCCL \cite{Pu2023DynamicCC} &96.3&96.5&96.9&75.3&76.8&70.2 &80.5&90.5&76.2 \\
        
        OpenCon \cite{sun2022open} &90.4&89.3&91.1
        &73.4&74.2&72.6
        &84.3&90.7&81.1 \\
        
        \midrule

        CLIP+ORCA\cite{Cao2021OpenWorldSL} &54.1 &74.5 &75.3 &36.3 &51.1 &34.1 &50.8 &58.2 &37.6 \\

        CLIP+GCD\cite{Vaze2022GeneralizedCD} &70.3& 92.1& 60.2 &50.7& 56.5& 39.0&55.3 &74.6 &45.6 \\

        CLIP+Promptcal\cite{promptcal} & 75.7& 94.3& 67.2& 56.8& 61.9& 43.6& 60.6&77.5 &49.3 \\

        CLIP-GCD\cite{clipgcd} &96.6 &97.2 &96.4 &85.2 &85.0 &85.6 &84.0 &85.0 &78.2\\

        Baseline-I &89.7 
        &90.6 &88.9 &67.7 &70.2 &57.5 &78.6 &88.2 &69.0 \\
        
        Baseline-II&91.2 &93.5&90.2 &76.2 &78.7 &64.5 
        &81.2 &90.3 &71.3 \\

        \textbf{GraphVL (ours)} &\textbf{97.9} &\textbf{98.5} &\textbf{98.3} &\textbf{87.6} &\textbf{87.1} &\textbf{87.8} &\textbf{86.6} &\textbf{95.9} 
        &\textbf{81.8} \\     
\bottomrule
        
    \end{tabular}}
    \label{tab1}
    \vspace{-0.2cm}
\end{table*}
\noindent\textbf{iii) Contextual Similarity Loss:} Although, the GCN module in our proposed GraphVL learns the semantic embeddings from the textual features in the supervision of respective visual features, but not sufficient enough to contextual information between the graph embeddings and learned textual prompts. To achieve this, we use the contextual similarity loss $\mathcal{L}_{CS}$, that computes the euclidean distance between the GCN embeddings and learned prompts for each class and helps in reducing the distance of prompt features to their respective centers.
\begin{equation}
   \mathcal{L}_{CS} = \frac{1}{2} \sum_{i=1}^{N} \|t_i - \mu_{\mathcal{G}(f_t(y_i))}\|^2
   \label{eq_csl}
\end{equation}
where, $\mu_{\mathcal{G}(f_t(y_i))}$ is the center of the GCN embedding for the class.\\
\noindent\textbf{Total Loss:} We optimize GraphVL using total loss $\mathcal{L}_{Tot}$, which consists of  losses defined in Equations \ref{eq_unssl}-\ref{eq_csl}. Also, we consider the unity weight among the loss terms for our experiments.
\vspace{-0.15cm}
\begin{equation}
    \mathcal{L}_{Tot} = \mathcal{L}_{CMA} + \mathcal{L}_{SDP} + \mathcal{L}_{CS}
    \label{eq_tot}
\end{equation}


\subsection{Semi-supervised clustering using $\mathbb{D_S} \cup \mathbb{D_U}$:} With the trained projectors ($P_v$ and $\mathcal{G}$) in our proposed GraphVL, we aim to group the samples in $\mathbb{D_U}$ into $|{Y_S}| + |{Y}_U|$ clusters while incorporating the labeled data from $\mathbb{D_S}$. The clustering process is semi-supervised, as we enforce label consistency for the samples from $\mathbb{D_S}$ during clustering—e.g., labeled samples from similar known classes should share cluster indices.

In contrast to the approach proposed in \cite{Vaze2022GeneralizedCD}, where clustering is performed in a non-semantic space, we propose leveraging the rich semantic structure learned by $\mathcal{G}$. Specifically, we represent the samples based on their similarity distributions with the class embeddings obtained from $\mathcal{G}$ for the $|{Y_S}|$ labeled classes. For each $x \in \mathbb{D_S} \cup \mathbb{D_U}$, we define the clustering feature of $x$ as $\mathcal{Q}(x) = [\delta(P_v(x), \bar{y}_1), \cdots, \delta(P_v(x), \bar{y}_{|{Y_S}|})]$. Since the semantic space is discriminative, samples from known classes will be closely associated with their respective class embeddings, while samples from novel classes will be projected into the space spanned by the known classes, resulting in distinctive clusters (Fig. \ref{fig:main_block}).

\section{Experimental Evaluations}
\noindent \textbf{Dataset descriptions:}
We evaluated GraphVL across seven diverse datasets: three general datasets (ImageNet-100 \cite{Krizhevsky2012ImageNetCW}, CIFAR-10 \cite{Krizhevsky2009LearningML}, CIFAR-100 \cite{Krizhevsky2009LearningML}), three fine-grained datasets (CUB-200 \cite{Wah2011TheCB}, StanfordCars \cite{Krause20133DOR}, Aircraft \cite{Maji2013FineGrainedVC}), and one granular dataset (iNaturalist \cite{iNaturalist}). ImageNet-100 contains over 14 million images in 21,841 subcategories. CIFAR-10 and CIFAR-100 consist of 32x32 images with 10 and 100 classes, respectively. CUB-200 has 11,788 bird images across 200 classes, StanfordCars includes 16,185 car images in 196 classes, and Aircraft comprises 10,200 images of 102 aircraft models. iNaturalist features 680,259 fine-grained images in 13 classes. For CIFAR-100, we used an 80/20 class split for known and novel classes, and a 50/50 split for the remaining datasets.\\
\noindent \textbf{Implementation details:}
We use the pre-trained CLIP ViT-L/14@336 \cite{vit} vision encoder ($f_v$) as the backbone, however, to enhance the feature representation of images, we apply a two layer trainable non-linear projection head ($P_v$) to the output of the vision projector. Furthermore, for text-based representations, we integrate a Transformer \cite{transformer} based text encoder ($f_t$) with a two-layer Graph Convolutional Network.
\cite{Kampffmeyer2018RethinkingKG}.
Additionally, we train our experiments for 100 epochs with a batch size of 128, an initial learning rate of $10^{-3}$, and the Adam optimizer \cite{kingma2014adam}, respectively.\\
\noindent \textbf{Evaluation metrics:}
Following \cite{Vaze2022GeneralizedCD}, we report: \textsc{All}: the clustering accuracy on all the samples in $\mathbb{D}_{U}$, \textsc{Known}: the clustering performance on the samples from $\mathbb{D}_{U}$ for the classes from ${C}_{kwn}$, and \textsc{New}: the clustering performance on the samples from $\mathbb{D}_{U}$ pertaining to the classes from ${C}_{new}$, respectively. We report the average performance over three runs.

\vspace{-0.5cm}
\subsection{Baselines}
We evaluate our GraphVL against benchmark methods, aiming to extend the capabilities of CLIP encoders. To address data leakage challenges associated with the CLIP, we present results in two ways. The first seven rows in Table \ref{tab1} and \ref{tab2} compare our model with the latest GCD literature. The subsequent rows show comparisons with a CLIP-reinforced GCD literature, using a CLIP encoder similar to ours. This fair assessment mitigates data leakage concerns. We use pre-trained CLIP features with K-means clustering as our baseline and compare with GCD/NCD techniques from recent literature \cite{Vaze2022GeneralizedCD, promptcal, Cao2021OpenWorldSL,Zhao2023LearningSG, Pu2023DynamicCC,sun2022open}, replacing their original feature backbones with the pre-trained CLIP's vision encoder for consistency. In addition, we consider CLIP-GCD\cite{clipgcd} only CLIP-based baseline for comparison as well.

We present two variants of our model to establish GraphVL's superiority. In \textbf{Baseline-I}, we replace the GCN layers with fully-connected layers (CLIP-GCD) to showcase the effectiveness of the Graph Convolutional Network module. This variant serves as a yardstick to quantify the direct impact of GCN in enhancing performance. In \textbf{Baseline-II}, we introduce a non-semantic clustering baseline, shifting focus from $\mathcal{Q}$ to the non-semantic space, emphasizing the importance of our similarity distributions as features. These baselines provide foundational context for evaluating the advancements and contributions of our GraphVL approach, assessing the effectiveness of the Graph Convolutional Network module and the significance of the proposed similarity distributions as features within our framework.

\begin{table*}[ht!]
    \centering
    \caption{Performance comparison to the state-of-the-art on the GCD task on the fine-grained and granular datasets. The first seven rows do not consider CLIP as a their feature backbone. The best accuracies are indicated in \textbf{bold}.}
    \vspace{-0.3cm}
    \scalebox{0.85}{
    \begin{tabular}{l|ccc|ccc|ccc|ccc}
    \toprule

       \multirow{3}{*}{\textbf{Method}}&\multicolumn{9}{c|}{\textbf{Fine-grained Datasets}}&\multicolumn{3}{c}{\textbf{Granular Dataset}}\\
       \cline{2-10} \cline{11-13}
        &\multicolumn{3}{c}{\textbf{CUB-200}}&\multicolumn{3}{c}{\textbf{StanfordCars}}&\multicolumn{3}{c|}{\textbf{Aircraft}}&\multicolumn{3}{c}{\textbf{iNaturalist}}\\
        \cline{2-4} \cline{5-7}\cline{8-10}\cline{11-13}
        
       & \textsc{All} & \textsc{Known} & \textsc{New} & \textsc{All} & \textsc{Known} & \textsc{New}  & \textsc{All} & \textsc{Known} & \textsc{New} & \textsc{All} & \textsc{Known} & \textsc{New} \\ 
       
        \midrule
        KMeans\cite{kmeans} &34.3 & 38.9& 32.1 & 12.8 & 10.6& 13.8 &12.9 &  12.9&  12.8 &12.1 &11.3 &12.9\\
        
        RankStats+ &33.3&51.6&24.2&28.3 &61.8&12.1&27.9&55.8&12.8
        &29.4&45.4&13.3\\
        
        UNO+  &35.1&49.0&28.1&35.5&70.5&18.6&28.3&53.7&14.7 
        &28.8&43.9&13.7\\
        
        GCD \cite{Vaze2022GeneralizedCD}  &51.3&56.6&48.7&39.0&57.6&29.9&35.4&51.0&27.0
        &40.8&58.6&22.9\\
        
        GPC \cite{Zhao2023LearningSG} &52.0&55.5&47.5 &38.2&58.9&27.4 &43.3&40.7&44.8
        &42.0&48.5&35.4\\
        
        DCCL \cite{Pu2023DynamicCC} 
        &63.5&60.8&64.9 &43.1&55.7&36.2 &54.4&56.7&52.2
        &49.1&58.2&40.0\\
        
        OpenCon \cite{sun2022open}
        &54.7&63.8&52.1
        &49.1&78.6&32.7
        &55.1&62.4&47.7
        &50.0&60.9&39.1\\
        
        \midrule
        
        CLIP+ORCA\cite{Cao2021OpenWorldSL} &23.9 &43.1 &24.7 & 33.9& 39.9& 32.4 &21.5 &27.3  & 25.9 &36.8 &45.8 &27.8 \\

        CLIP+GCD\cite{Vaze2022GeneralizedCD} &39.2&41.0&	38.2  &46.5	&58.9&	40.3  &27.5	&27.7	&27.5  &{33.8} &48.2 &{29.3}   \\

        CLIP+Promptcal\cite{promptcal} &59.7 &57.3 &52.7 &65.6 &73.6 &58.3 &55.2 &57.7 &52.4 &{45.8} &50.0 &{41.6} \\

        CLIP-GCD\cite{clipgcd} &62.8 &77.1 &55.7 &70.6 &88.2 &62.2 &50.0 &56.6 &46.5 &- &- &-\\

        Baseline-I 
        &68.0 &86.3 &49.7 &65.7 &74.2 &57.4 &45.9 &47.4 &44.5
        &50.2 &59.9 &40.6\\
        
        Baseline-II &71.4 &88.4 &56.3 &68.2 &76.6 &61.0 &52.6 &59.1 &48.8
        &62.6 &61.3 &63.9\\

        \textbf{GraphVL (ours)} &\textbf{78.5} &\textbf{94.1} &\textbf{75.8} &\textbf{81.5} &\textbf{87.3} &\textbf{73.6} &\textbf{64.2}  &\textbf{65.6} &\textbf{55.9} &\textbf{65.1} &\textbf{64.4} &\textbf{65.8}\\       
\bottomrule
        
    \end{tabular}}
    \label{tab2}
    \vspace{-0.3cm}
\end{table*}

\subsection{Comparison to the literature} 
In Table \ref{tab1}, GraphVL is rigorously evaluated on popular image classification datasets, outperform SOTA methods such as CLIP-adapted ORCA, GCD, and PromptCAL. Across all datasets, GraphVL demonstrates superior performance, surpassing the baselines at least by $1.4\%$ on CIFAR-10, $2.2\%$ on CIFAR-100, and $0.7\%$ on ImageNet-100, particularly excelling on new classes. Notably, GraphVL showcases a significant advantage over CLIP-GCD, emphasizing its effectiveness in learning a topology-preserving and discriminative semantic mapping. Furthermore, evaluations on fine-grained datasets in Table \ref{tab2} highlights GraphVL's superiority over CLIP-adapted ORCA, GCD, and PromptCAL, outperforming them across all classes. Notably, GraphVL achieves substantial performance gains, exceeding the baselines at least by $10.9\%$ in CUB-200, $11.4\%$ in StanfordCars, and $3.5\%$ in Aircraft datasets on new classes. These results establish GraphVL as the new state-of-the-art architecture in the field, consistently outperforming CLIP-GCD. 

Finally, we evaluate our GraphVL on the challenging iNaturalist dataset, and our proposed GraphVL surpasses the existing methods with a substantial margin of $2.5\%$ across all categories, $3.1\%$ for known classes, and $1.9\%$ for new classes when using the CLIP-based visual backbone. These findings underscore GraphVL's excellence, especially on datasets with intricate granularity.
\subsection{Model Ablation and Critical analysis}
\begin{table}[ht!]
    \centering
    \caption{Ablation analysis of our proposed GraphVL for number of GCN layers for text-embeddings.}
    \vspace{-0.3cm}
    \scalebox{0.9}{
    \begin{tabular}{c|ccc|ccc}
    \toprule
        \multirow{2}{*}{\textbf{No. of GCN Layers}}&\multicolumn{3}{c|}{\textbf{CIFAR-10}}&\multicolumn{3}{c}{\textbf{Aircraft}}\\
        \cline{2-4} \cline{5-7}
        
       & \textsc{All} & \textsc{Known} & \textsc{New} & \textsc{All} & \textsc{Known} & \textsc{New}   \\  
        \midrule
        0 &92.6 &97.7 &85.8 &50.5 &52.5 &48.5  \\
        1 &94.6 &93.1 &96.1 &51.4 &51.5 &51.2  \\
        2 &\textbf{97.9} &\textbf{98.5} &\textbf{98.3} &\textbf{64.2} &\textbf{65.6} &\textbf{55.9}  \\
        3 &94.9 &94.0 &95.9 &55.1 &58.9 &51.4 \\
\bottomrule
\end{tabular}}
    \label{abl_tab1}
    \vspace{-.42cm}
\end{table}
\begin{figure}[ht!]
    \centering
    \includegraphics[width=0.95\columnwidth]{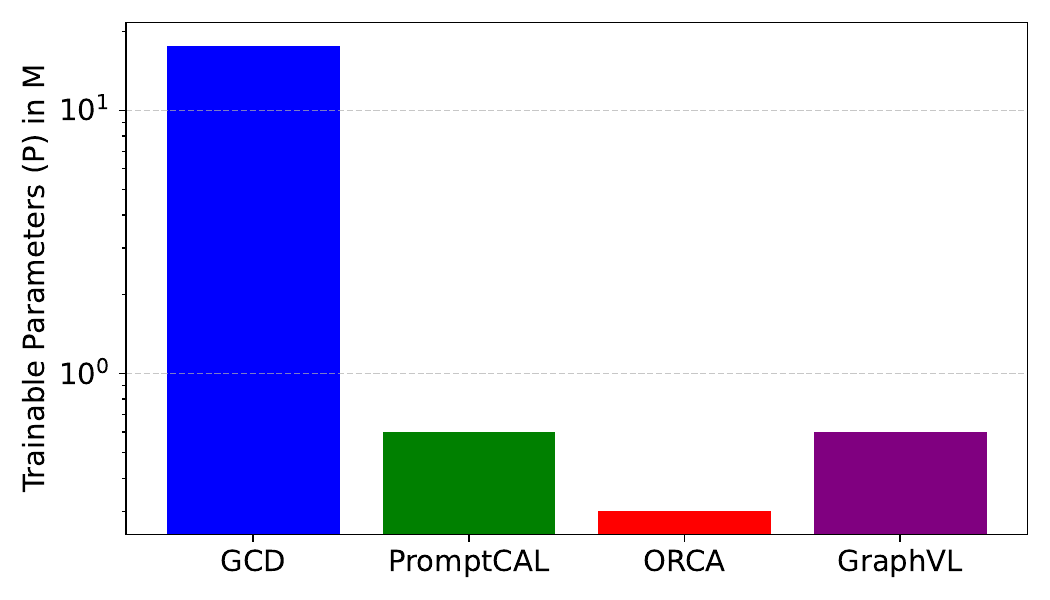}
    \vspace{-5mm}
    \caption{Comparing the number of trainable parameters in GraphVL and state-of-the-art class discovery methods.}
    \vspace{-7mm}
    \label{abl_tab6}
\end{figure}
\begin{figure*}[ht!]
\centering

\includegraphics[width=0.8\columnwidth]{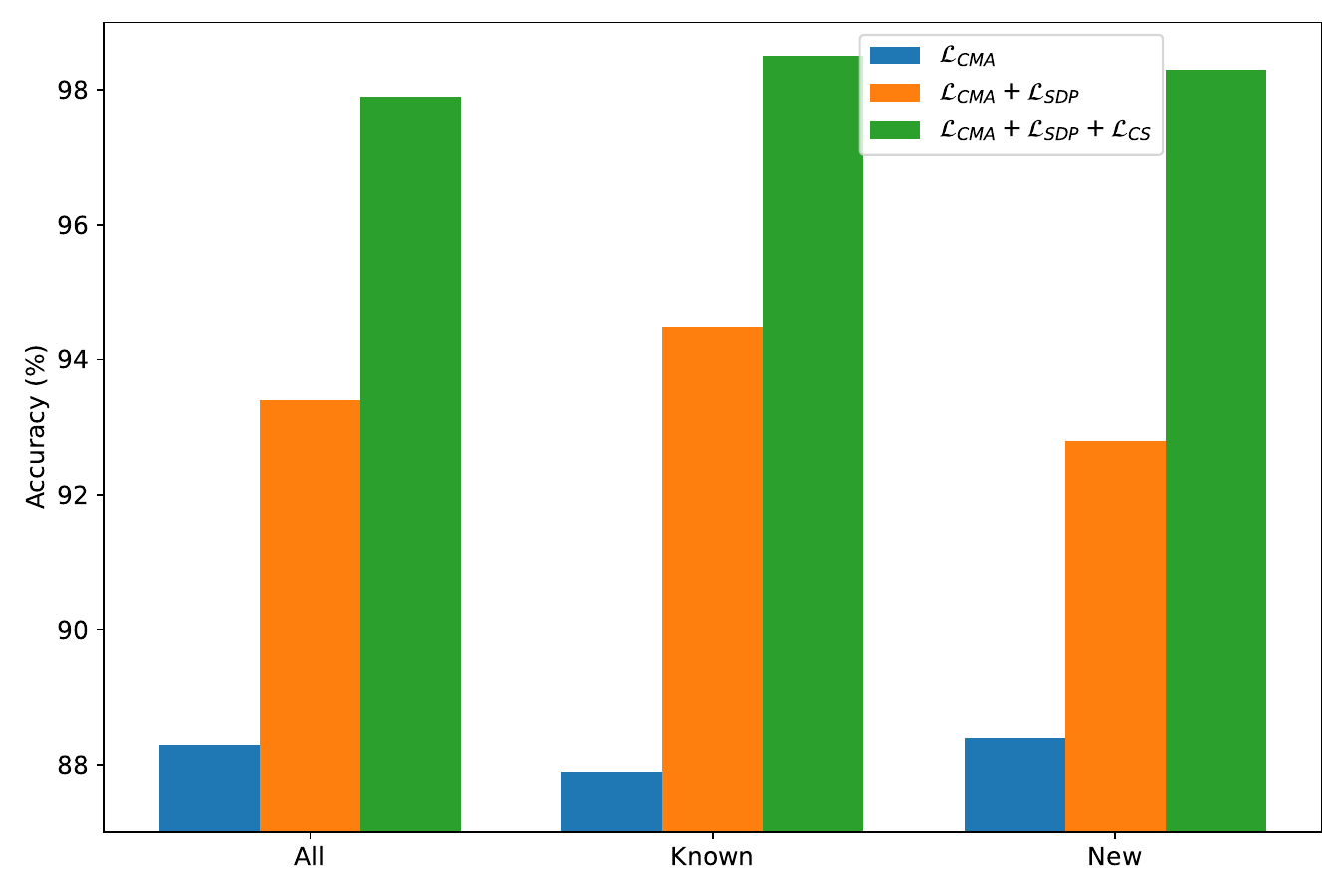}
\includegraphics[width=0.8\columnwidth]{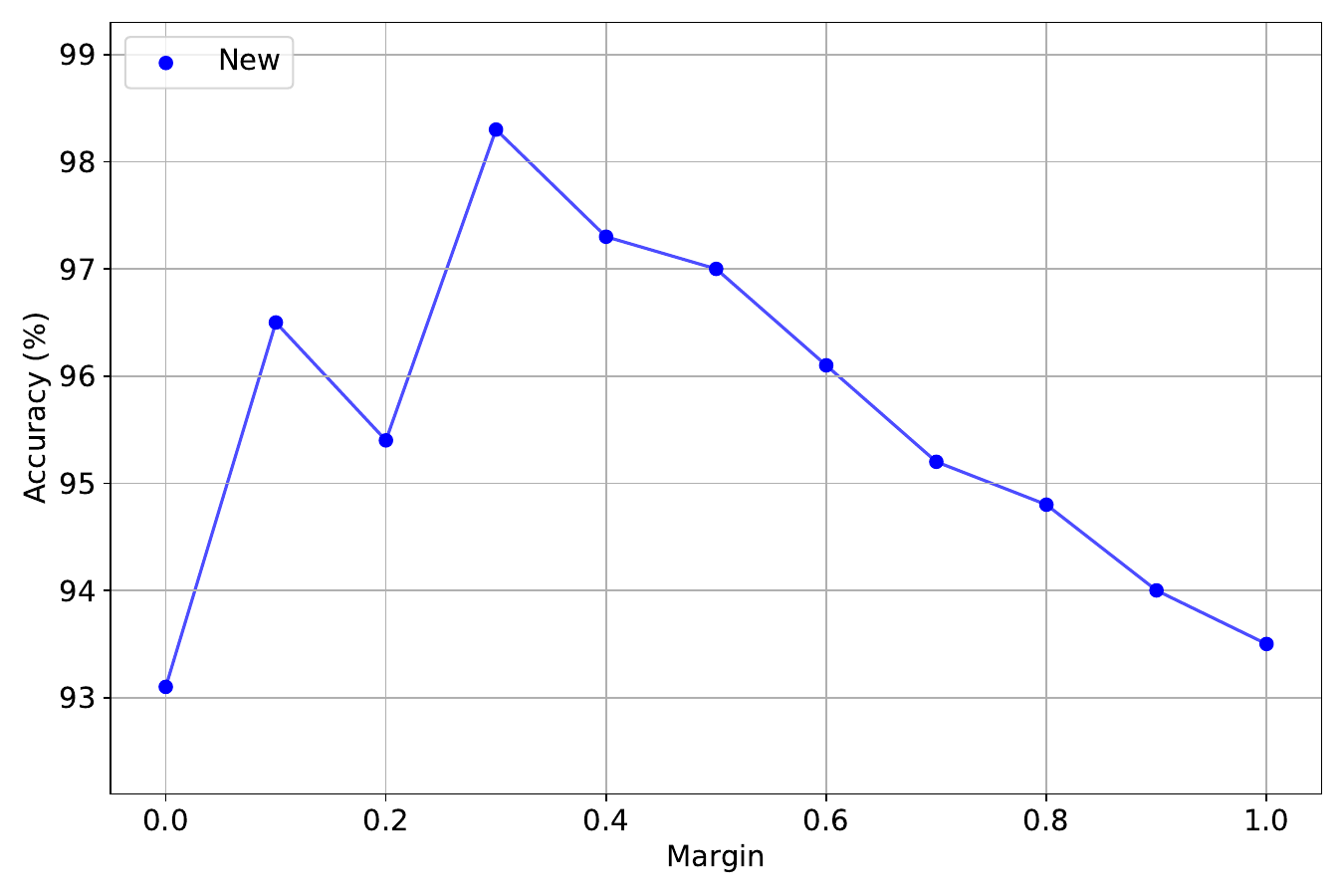}
\vspace{-0.5cm}
\caption{The sensitivity analysis of the loss terms and alpha using bar graph (\textbf{left}) and line graph (\textbf{right}), respectively.}\label{ablation}
\vspace{-2mm}
\end{figure*}
\noindent \textbf{Sensitivity to GCN Depth:}
Here, we investigate the effectiveness of Graph Convolutional Networks (GCNs) for discovering new classes, as detailed in Table \ref{abl_tab1}. Specifically, we're exploring how the number of layers in a GCN affects its ability to understand hidden characteristics of unfamiliar classes, all the while avoiding the trap of overemphasizing previously known classes. Our experiments have led us to a noteworthy observation. It appears that GCNs function optimally when structured with two layers for message-passing. This configuration strikes a balance, allowing the GCN to adeptly learn from both new and existing classes, resulting in better performance compared to GCNs with just one layer or three layers. In essence, our investigation underscores the importance of a well-judged two-layer configuration in GCNs. This empirical finding, backed by thorough experimentation, highlights its superiority in grasping the nuances of emerging class features, while simultaneously sidestepping the pitfalls of excessive adaptation to well-established categories.\\
\noindent \textbf{Comparison of the model complexity:} 
GraphVL showcases superior parameter efficiency, outperforming GCD \cite{Vaze2022GeneralizedCD} by a fair margin. Despite having more parameters than PromptCAL \cite{promptcal}, GraphVL impressively reduces its count by about 29 times compared to GCD (Table \ref{abl_tab6}). Notably, it surpasses PromptCAL's parameter tally by 0.93 times. While GraphVL has more parameters than ORCA \cite{Cao2021OpenWorldSL}, its exceptional performance across all datasets sets it apart, demonstrating a substantial leap in effectiveness. This accomplishment underscores the significant performance improvement of GraphVL compared to its competitors.\\

\begin{table*}[ht!]
    \centering
    \caption{Performance comparision of different vision  backbones of the pre-trained CLIP.}
    \vspace{-0.2cm}
    \scalebox{0.85}{
    \begin{tabular}{l|ccc|ccc|ccc|ccc}
    \toprule
        \multirow{2}{*}{\textbf{Method}}&\multicolumn{3}{c|}{\textbf{CIFAR10 \cite{Krizhevsky2009LearningML}}}&\multicolumn{3}{c|}{\textbf{CIFAR100 \cite{Krizhevsky2009LearningML}}}&\multicolumn{3}{c|}{\textbf{Aircraft \cite{Maji2013FineGrainedVC}}}
        &\multicolumn{3}{c}{\textbf{CUB-200 \cite{Wah2011TheCB}}}\\
        \cmidrule(lr){2-4} \cmidrule(lr){5-7}\cmidrule(lr){8-10}\cmidrule(lr){11-13}
        
       & \textsc{All} & \textsc{Known} & \textsc{New} & \textsc{All} & \textsc{Known} & \textsc{New}  & \textsc{All} & \textsc{Known} & \textsc{New}  & \textsc{All} & \textsc{Known} & \textsc{New} \\ 
        \midrule

        RN-50
        &71.05 &80.96 &61.14 &63.10 &67.23 &41.70 &24.63 &22.13 &27.14 &59.21 &73.32 &54.43 \\

        ViT-B/16
        &90.69 &90.11 &91.28 &72.43 &75.09 &66.35 &37.22 &34.84 &39.59 &70.72 &82.84 &64.59 \\

        ViT-B/32
         &90.36 &93.03 &88.22 &71.91 &75.91 &63.53 &43.26 &47.59 &38.93 &72.43 &85.84 &69.59 \\
      
        \textsc{\textbf{ViT-L/14}} &\textbf{97.90} &\textbf{98.52} &\textbf{98.34} &\textbf{87.61} &\textbf{87.10} &\textbf{87.79} &\textbf{64.19} &\textbf{65.60} &\textbf{55.91} &\textbf{78.50} &\textbf{94.07} &\textbf{75.76} \\

\bottomrule
        
    \end{tabular}}
    \label{abl_tab3}
    \vspace{-0.3cm}
\end{table*}
\noindent \textbf{Sensitivity to the Loss Terms and $\alpha$ Margin:} In Fig. \ref{ablation} (left), we investigate the impact of various components of our loss function. Specifically, we analyze the contrastive loss $\mathcal{L}_{CMA}$, both individually and in combination with $\mathcal{L}_{SDP}$ and the cumulative $\mathcal{L}_{CS}$. Our results reveal that combining all three loss components yields the best outcomes. We find that $\mathcal{L}_{CMA}$ alone is insufficient to surpass baseline performance, achieving only 88.3\% accuracy on the novel classes of the CIFAR-10 dataset. In contrast, combining $\mathcal{L}_{CMA}$ with $\mathcal{L}_{SDP}$ results in a notable 5\% increase in new accuracy compared to $\mathcal{L}_{CMA}$ alone, underscoring the crucial role of both $\mathcal{L}_{CMA}$ and $\mathcal{L}_{SDP}$ in enhancing object recognition. However, even this combination does not match the performance of our baseline CLIP-GCD. Incorporating $\mathcal{L}_{CS}$ further improves performance. The combination of all three losses achieves the highest accuracy of 98.3\%. These insights emphasize the importance of each loss term in achieving superior performance in recognizing novel classes.

In Fig. \ref{ablation} (right), we examine the sensitivity of our loss $\mathcal{L}_{CMA}$ to the margin parameter $\alpha$. We assess margins ranging from 0 to 1 and find optimal performance when the margin is set to $0.3$. This finding is consistent across various datasets, including CIFAR-10. Additionally, t-SNE \cite{van2008visualizing} plots for $\alpha = 0.3$ in Fig. \ref{fig:tsne} show significant separation between known and unknown classes.

\noindent \textbf{Sensitivity to different feature backbones:} In Table \ref{abl_tab3}, we've taken a deep dive into evaluating the performance of different CLIP backbones in the context of uncovering new classes within the CIFAR-10 and Aircraft datasets. Our experiments have illuminated an interesting insight: the CLIP model equipped with the memory-intensive CLIP ViTL/14 backbone reigns as the most potent contender. This particular configuration surpasses its counterparts, namely CLIP B/16, CLIP B/32, and RN-50, emerging as the champion in our trials.  In essence, our findings underscore the superior prowess of the CLIP model outfitted with the CLIP ViTL/14 backbone. This shines a light on its exceptional capability in outshining other variants, including those with less memory-intensive backbones, when it comes to unearthing novel classes across the datasets under consideration.
\begin{table}[ht!]
     \vspace*{-2mm}
      \caption{Ablation of our proposed GraphVL with linear layer vs GCN layer for encoding the textual prompts.}
    \vspace{-0.2cm}
    \centering
    \scalebox{0.8}{
    \begin{tabular}{c|ccc|ccc}
    \toprule
        \textbf{No. of Linear Layers}&\multicolumn{3}{c|}{\textbf{CIFAR-10}}&\multicolumn{3}{c}{\textbf{Aircraft}}\\
        \cline{2-4} \cline{5-7}
        
       & \textsc{All} & \textsc{Known} & \textsc{New} & \textsc{All} & \textsc{Known} & \textsc{New}   \\  
        \midrule
        1&89.7 &90.6 &88.9& 45.9& 47.4& 44.5\\
        2 &{93.2} &97.8&{88.3} &{48.2} &{49.2} &{47.2}  \\
         \textbf{with GCN Layers (ours)}&\textbf{{97.9}} &\textbf{98.5} &\textbf{{98.3}} &\textbf{{64.2}} &\textbf{{65.6}} &\textbf{{55.9}}  \\
    \bottomrule
    \end{tabular}}
    \label{tab:abl1}
\end{table}

    
\begin{table}[ht!]
 \caption{The estimated number of clusters by GraphVL with the original number of classes (ground truth as G.T.).}
    \vspace{-0.3cm}
    \centering \scalebox{0.8}{
    \begin{tabular}{l|c|c|c}
    \midrule
        \textbf{Datasets} & \textbf{G.T.} & \textbf{GraphVL} & \textbf{Error (\%)} \\
      \midrule
        \textbf{CIFAR-10} & 10 & 10 & 0.0 \\
        \textbf{CIFAR-100} & 100 & 97 & 3.0 \\
        \textbf{Imagenet} & 100 & 95 & 5.0 \\
        \textbf{CUB-200} & 200 & 199 & 0.5 \\
        \textbf{StanfordCars} & 196 & 177 & 9.6 \\
        \textbf{Aircraft} & 100 & 84 & 16.0 \\
        \bottomrule
    \end{tabular}}
    
    \label{cluster}
    \vspace{-0.4cm}
\end{table}

\begin{figure*}[h!]\centering
    \includegraphics[width=0.55\columnwidth]{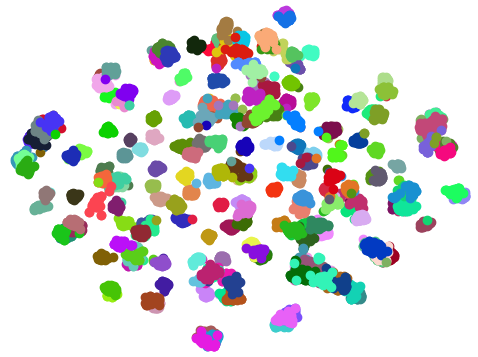}
    \includegraphics[width=0.65\columnwidth]{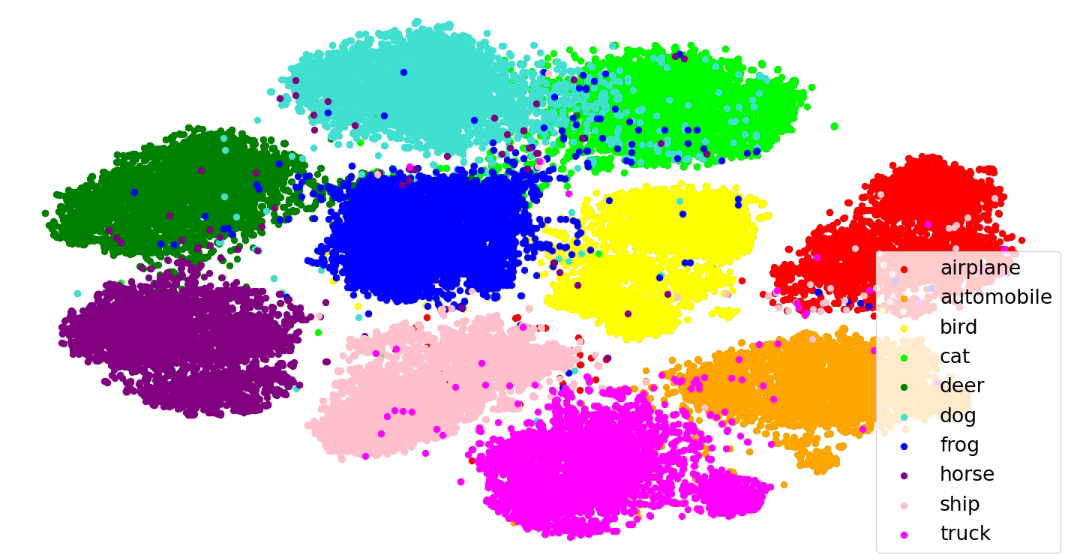}
\vspace{-0.25cm}
    \caption{t-SNE plots clustering the classes in StanfordCars (left) and CIFAR-10 (right) using our proposed GraphVL.}
    \vspace{-2mm}
    \label{fig:tsne}
\end{figure*}

\noindent \textbf{Advantage of GCN over linear layers:} 
Table \ref{tab:abl1} demonstrates the impact of replacing the information passing GCN with linear layers and non-linear activations in GraphVL. It becomes evident that linear layers demonstrate a commendable proficiency in effectively encapsulating well-established data patterns; however, their performance notably declines when dealing with the intricate features of unfamiliar samples.

\noindent\textbf{t-SNE visualization and estimating the number of clusters:} By employing t-SNE \cite{van2008visualizing}, we unravel complex dataset structures. In CIFAR-10 (Fig. \ref{fig:tsne} left), distinct clusters represent high-level object categories. The StanfordCars dataset (Fig. \ref{fig:tsne} right) showcases t-SNE's prowess in capturing subtle distinctions, enriching our insights into fine-grained data aspects.
\begin{figure*}[h!]\centering
    \includegraphics[scale=0.32]{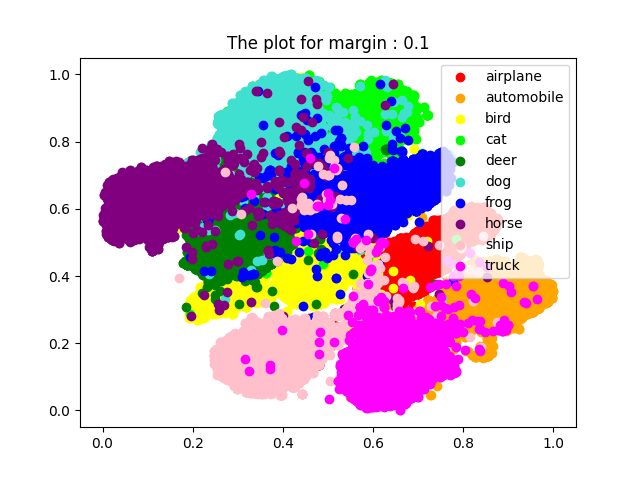}
    \includegraphics[scale=0.32]{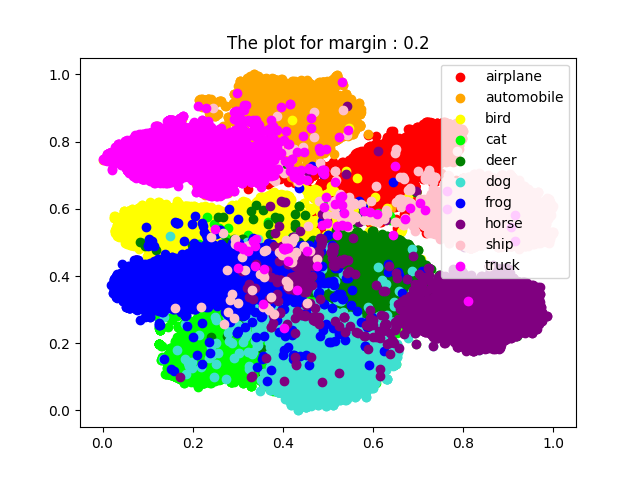}
    \includegraphics[scale=0.32]{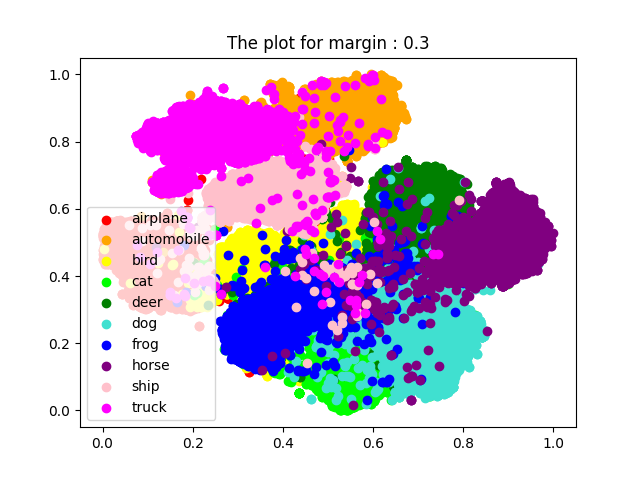}
    \includegraphics[scale=0.32]{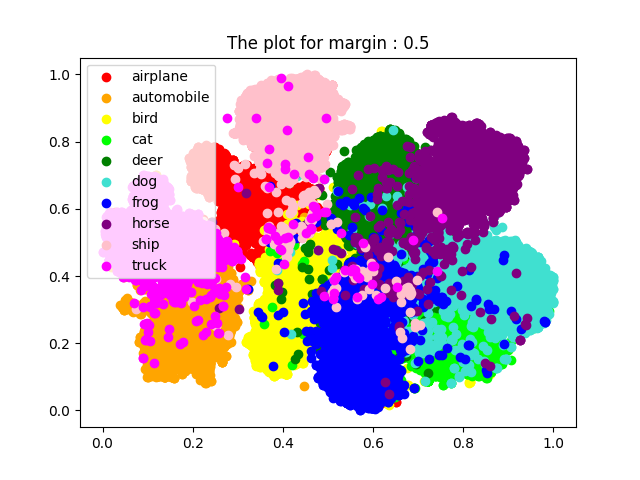}
    \includegraphics[scale=0.32]{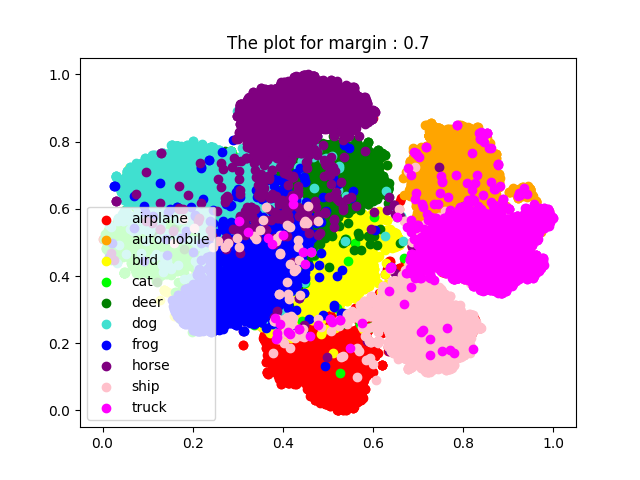}
    \includegraphics[scale=0.32]{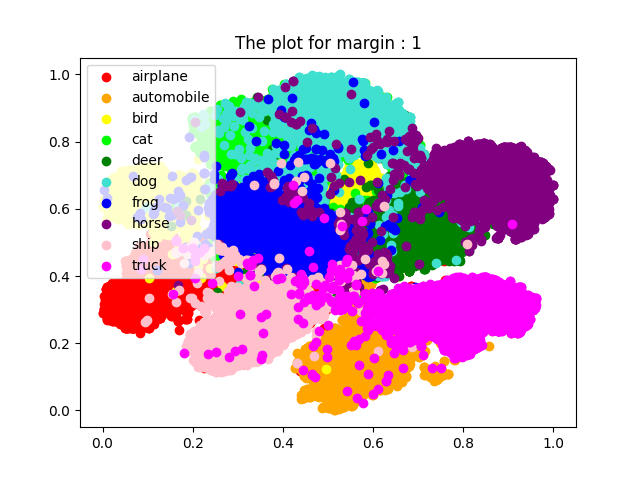}
    \vspace{-0.4cm}
    \caption{T-SNE plots between the image feature extracted from the GraphVL for different values of margin in the $\mathcal{L}_{CMA}$ loss.}
    \label{fig:tsne}
    \vspace{-3.5mm}
\end{figure*}

In Fig. \ref{fig:tsne}, we present a t-SNE visualization of the image embeddings generated by GraphVL for different margin values of the loss $\mathcal{L}_{CMA}$. The t-SNE visualization technique offers valuable insights into the impact of different margin values on class separation within deep learning models. By plotting t-SNE visualizations for various margin values, we gain a visual representation of the distribution and separation of known and unknown classes. Notably, our analysis reveals that lower margin values, such as 0.1 and 0.2, highlight substantial separation among the classes, resulting in distinct clusters. As the margin value increases, the separation diminishes, leading to overlapping clusters and reduced discrimination between classes. However, with a margin value of 0.3, we observe an optimal balance between class separation and overlap, as the t-SNE plots exhibit well-defined clusters with minimal inter-class overlap. These findings are consistent across the CIFAR10 dataset and hold true for other datasets as well, reinforcing the significance of margin value selection in achieving effective class separation and model performance in deep learning tasks.

In the context of class number estimation, our results are summarized in Table \ref{cluster}. We employ the elbow method, a well-established technique in unsupervised machine learning, to estimate the optimal number of clusters in a dataset. This method involves iteratively adjusting the value of K, which represents the number of clusters, and assessing the corresponding within-cluster variation. The point of inflection, often referred to as the `elbow', signifies a trade-off between cluster compactness and fragmentation, serving as a heuristic for determining the cluster count. For general object recognition datasets, our method approximates the true number of categories in the unlabeled set with minimal error. Our approach deviates by only 3.33\% from the average ground truth classification count. On fine-grained datasets, we observe an average estimation error of 8.70\%. This nuanced difference highlights the complexities of datasets with visually similar classes.

\section{Takeaways and Future Avenues}
We have introduced a novel approach to tackle the GCD problem by harnessing the power of the vision-language model (CLIP). Our primary focus is on leveraging the semantic knowledge embedded in CLIP's representation space to enhance the clustering performance on unlabeled data. To address the bias issue in GCD, we employ a learnable GCN to preserve neighborhood information in the semantic space, along with a pair of metric losses in the non-semantic space and visual-semantic mapping for discriminability. For clustering the unlabeled data using semi-supervised k-means, we propose to utilize the discriminative semantic similarity distribution as features. Our experimental results on seven datasets demonstrate the efficay of our GraphVL approach compared to existing literature. Currently, we are exploring avenues to scale our model for handling a larger number of novel classes.

\bibliographystyle{ACM-Reference-Format}
\bibliography{cite}





\end{document}